\documentclass[sigconf, nonacm]{acmart}

\usepackage{amsmath}
\usepackage{algorithmic}
\usepackage{algorithm}
\usepackage{multirow}
\usepackage{subfig}
\usepackage{graphicx}
\usepackage{xcolor}
\usepackage{eqparbox}

\AtBeginDocument{%
  \providecommand\BibTeX{{%
    \normalfont B\kern-0.5em{\scshape i\kern-0.25em b}\kern-0.8em\TeX}}}

\setcopyright{acmcopyright}
\copyrightyear{2023}
\acmYear{2023}
\acmDOI{XXXXXXX.XXXXXXX}

\acmConference[KDD '23]{29th ACM SIGKDD Conference on Knowledge Discovery and Data Mining}{Aug 06--10,
  2023}{Long Beach,California}
\acmPrice{15.00}
\acmISBN{978-1-4503-XXXX-X/18/06}

\begin{document}

\title{Learning Invariant Rules from Data for Interpretable Anomaly Detection}
\author{Cheng Feng}
\email{cheng.feng@siemens.com}
\affiliation{%
  \institution{Siemens AG}
  \city{Beijing}
  \country{China}
}

\author{Pingge Hu}
\email{hpg18@mails.tsinghua.edu.cn}
\affiliation{%
  \institution{Tsinghua University}
  \city{Beijing}
  \country{China}
}


\begin{abstract}
In the research area of anomaly detection, novel and promising methods are frequently developed. However, most existing studies exclusively focus on the detection task only and ignore the interpretability of the underlying models as well as their detection results. Nevertheless, anomaly interpretation, which aims to provide explanation of why specific data instances are identified as anomalies, is an equally important task in many real-world applications. In this work, we propose a novel framework which synergizes several machine learning and data mining techniques to automatically learn invariant rules that are consistently satisfied in a given dataset. The learned invariant rules can provide explicit explanation of anomaly detection results in the inference phase and thus are extremely useful for subsequent decision-making regarding reported anomalies. Furthermore, our empirical evaluation shows that the proposed method can also achieve comparable or even better performance in terms of AUC and partial AUC on public benchmark datasets across various application domains compared with start-of-the-art anomaly detection models.
\end{abstract}



\maketitle

\section{Introduction}
\label{sec:introduction}
Anomaly detection, which aims to identify data instances that do not conform to the expected behavior, is a classic data mining task with numerous applications in process condition monitoring, network intrusion detection, equipment fault detection, etc. Over the past decades, numerous methods have been proposed to tackle this challenging problem in different application domains. Examples include one-class classification-based \cite{manevitz2001one,tax2004support,ruff2018deep}, nearest neighbor-based \cite{breunig2000lof,tang2002enhancing}, clustering-based \cite{yu2002findout,jiang2008clustering}, isolation-based \cite{liu2008isolation,liu2012isolation,hariri2019extended}, density-based \cite{rousseeuw1999fast,kim2012robust,feng2021time,liu2022unsupervised} and deep anomaly detection models based on autoencoders \cite{chen2017outlier,zhou2017anomaly,zong2018deep}, generative adversarial networks \cite{zenati2018adversarially,han2021gan}, to name a few. We refer to \cite{chandola2009anomaly,pang2021deep} for a comprehensive review. However, we realize that most existing studies exclusively focus on the detection task only and ignore the interpretability of the underlying models and their detection results. In fact, anomaly interpretation, which aims to provide explanation of why specific data instances are detected as anomalies, is equally critical for many real-world applications. For instance, when an anomaly is reported by a fault detection application for a critical device in a factory, human experts need straightforward clues regarding the reported anomaly, and then can decide what next steps - such as fault diagnosis, predictive maintenance and system shutdown – should be taken. Examples of required clues include which signal(s) is abnormal, why the anomaly detector thinks that signal(s) is abnormal and what is the supposed value for that abnormal signal(s). In this work, we propose an interpretable anomaly detection algorithm that can explicitly answer the above three critical questions.

Concretely, we present a novel method which synergizes several machine learning and data mining techniques including decision tree learning and association rule mining to learn invariant rules that are consistently satisfied in a given dataset. The learned invariant rules can not only provide accurate anomaly detection but also highly understandable explanations about reported anomalies in the inference phase. While significantly improving the interpretability of anomaly detection tasks, experimental results on benchmark datasets in different application domains show that our method can also achieve comparable or even better performance in terms of Area Under the Receiver Operating Characteristic Curve (AUC ROC) and partial AUC \cite{mcclish1989analyzing} compared with various start-of-the-art anomaly detection models. The data and code of our experiments are published in a GitHub repository\footnote{https://github.com/NSIBF/InvariantRuleAD} for better reproducibility of our results.

The rest of this paper is structured as follows. We briefly introduce association rule mining and decision tree learning which are the main techniques that are utilized in our method in the next section. In Section~\ref{sec:def} the formal definition of invariant rules is presented. Section~\ref{sec:learning} presents the method for learning invariant rules from training data. In Section~\ref{sec:inference}, we introduce how to use the learned invariant rules for anomaly detection and interpretation in the inference phase. Experiments for demonstrating the anomaly detection performance and interpretability of our proposed method are given in Section~\ref{sec:experiments}. Related work is discussed in Section~\ref{sec:related}. Lastly, Section~\ref{sec:conclude} draws final conclusions.

\section{Preliminaries}
\subsection{Association Rule Mining}
Association rule mining, one of the most important data mining techniques, is used to discover the frequently occurring patterns in the database \cite{agrawal1993mining}. The main aim of association rule mining is to find out the interesting relationships and correlations among the different items of the database. Specifically, let $I=\{i_1, i_2, \ldots i_m \}$ be a set of items and $D$ be a set of transactions, where each transaction $T$ is a set of items such that $T \subseteq I$. An association rule is expressed as $X \Rightarrow Y$ where $X,Y \subseteq I$ and $X \cap Y=\emptyset$. Furthermore, there are two basic measures for an association rule: support (s) and confidence (c). A rule has support $s$ if $s$ proportion of the transactions in $D$ contains $X \cup Y$. A rule $X \Rightarrow Y$ has confidence $c$, if $c$ proportion of transactions in $D$ that support X also support Y. Given a set of transactions $D$ (the database), the problem of mining association rules is to discover all association rules that have support and confidence greater than the user-specified minimum support (called \emph{minsup}) and minimum confidence (called \emph{minconf}). More specifically, association rules are commonly generated using the following two steps:
1) Find all the frequent itemsets whose support is larger than \emph{minsup}.
2) Based on these frequent itemsets, association rules which have confidence above \emph{minconf} are generated. The first step is much more difficult than the second step. Many algorithms have been developed to mine frequent itemsets, including Apriori \cite{agrawal1994fast}, FP-Growth \cite{han2000mining} and genetic algorithms \cite{sharma2012survey}, etc. 


In practice, the support of different items can vary significantly in the database, thus only using a single \emph{minsup} can cause problems such as dominating by infrequent items if $\emph{minsup}$ is too low or only covering a very limited items if $\emph{minsup}$ is too high. To solve the problems, different rules may need to satisfy different minimum supports depending on what items are in the rules, and we need to mine frequent itemsets with multiple minimum supports. This problem is much harder because the downward closure property (all non-empty subsets of a frequent itemset must also be frequent), which is the basis to reduce the search space of frequent itemsets in aforementioned algorithms, is invalid. Fortunately, there are available algorithms for mining frequent itemsets with multiple minimum support thresholds. Among the examples are MSApriori \cite{liu1999mining}, CFPgrowth \cite{hu2006mining} and CFP-growth++ \cite{kiran2011novel}. CFPgrowth and CFP-growth++ are more efficient than MSApriori when dealing with large databases.

\subsection{Decision Trees}
Decision trees (DTs) are a family of machine learning algorithms primarily designed for classification and regression. Their representability and ability to produce rules with relevant attributes make them the most commonly used technique when seeking interpretable machine learning models \cite{freitas2014comprehensible,nanfack2022constraint}. 

Specifically, let $X$ be the input variables with $M$ dimensions $X_i$ where $i=1,\ldots,M$, Y be the output variable ($Y \in \{1,\dots,C\}$ for classification, $Y \in \mathcal{R}$ for regression), $\mathcal{D}$ be the dataset formed by sampling from the unknown joint distribution $P_{XY}$. A DT consists of a hierarchy of internal nodes with defined splitting rules based on $X$, and a set of leaf nodes with predictions about $Y$. Splitting rules can involve one variable each time leading to univariate DTs, or multiple variables each time leading to multivariate DTs. To promote interpretability, we only consider univariate DTs in this work. Following \cite{nunes2020learning}, we denote a decision rule on a single variable $X_i$ as $f(\mathbf{x}) = \mathbf{1}_A(x_i)$ where $\mathbf{1}_A(x_i)$ is an indicator function, taking value 1 for $x_i \in A$ and 0 otherwise. For numerical $X_i$, we have:
$f(\mathbf{x}) = \mathbf{1}_{(\tau,\infty)}(x_i)$
where $\tau$ is the selected cut-off value. Intuitively, a 0/1 outcome directs the instance $\mathbf{x}$ to the left/right child node.

The learning of DTs is mainly composed by induction and pruning. Induction is about learning the DT structure and its splitting rules. \cite{laurent1976constructing} shows that learning an optimal DT that maximizes prediction performance while minimizing the size of the tree is NP-complete owing to the discrete and sequential nature of the splits. As a result, standard DT algorithms such as CHAID \cite{kass1980exploratory}, CART \cite{breiman2017classification}, ID3 \cite{quinlan1986induction}, and C4.5 \cite{quinlan2014c4} learn a DT by following locally optimal induction strategies. Specifically, locally optimal induction selects splitting rules that maximize an objective at each node, e.g., variance reduction for regression, maximizing information gain for classification, etc. The splitting procedure stops when a specific criterion, e.g., the maximum depth of the tree and the minimum number of samples required to be at a leaf node, is reached and a leaf node is created. By using this locally optimal search heuristic, learned greedy trees can be very accurate but significantly overfit. To avoid that, pruning techniques are commonly applied to find a trade-off between reducing the complexity of the tree and maintaining a certain level of accuracy \cite{barros2015automatic,nanfack2022constraint}.

\section{Definition of Invariant Rules}
\label{sec:def}
Let $X$ be a $M$ dimensional variable where each dimension $X_i \in \mathcal{R}$ is a continuous variable, $U$ be a $N$ dimensional variable where each dimension $U_i \in \{1,\ldots, C\}$ is a categorical variable. Given a training dataset $\mathcal{D} = ( \mathbf{d}^1,\mathbf{d}^2,\ldots,\mathbf{d}^{|\mathcal{D}|} )$ with no anomalies (e.g., collected from a controlled environment), and each data point\footnote{For convenience, we drop the superscript for sequential index when the context is clear} $\mathbf{d}=( x_1,\ldots,x_M, u_1,\ldots,u_N )$ is a vector consisting of a data instance for $X$ and $U$ respectively. Furthermore, we assume there exist a global set of predicates $\mathcal{P}=\{ p_1,  p_2, \ldots, p_{|\mathcal{P}|} \}$ (how the global predicate set is generated will be introduced in the subsequent section). For example, $5\leq X_i<10$ is a predicate for a continuous variable, $U_i=1$ is a predicate for a categorical variable. Each data point $\mathbf{d}$ may satisfy a subset of predicates in $\mathcal{P}$.

We further denote the support of a predicate set $\mathcal{S}$ ($\mathcal{S}\subseteq \mathcal{P}$) by $sup(S)$, representing the fraction of data points where all the predicates in $S$ are satisfied. Then, we formally define an invariant rule as follows:
\begin{eqnarray*} \label{eq:invariant}
\mathcal{S}_1 \Rightarrow \mathcal{S}_2 \texttt{ where }  \mathcal{S}_1,\mathcal{S}_2 \subseteq \mathcal{P} \texttt{ and }  \mathcal{S}_1 \cap \mathcal{S}_2=\emptyset 
\end{eqnarray*}where we call $\mathcal{S}_1$ the antecedent predicate set, $\mathcal{S}_2$ the consequent predicate set. The rule means whenever a data point satisfies all the predicates in $\mathcal{S}_1$, then it must also satisfy all the predicates in $\mathcal{S}_2$. For example, an invariant rule may look as follows:
\begin{eqnarray*}
5\leq X_1 <10, X_2>20.4, U_1=0  \Rightarrow  X_3<7.1, U_2 = 2  
\end{eqnarray*}
We also require any invariant rule must satisfy two conditions, namely the minimum support condition and $100\%$ confidence condition. Regarding the minimum support condition, we require the support of an invariant rule to be larger than a rule-specific minimum support threshold. Concretely, let $\mathcal{S}_1 \Rightarrow \mathcal{S}_2$ be an invariant rule, $\mathcal{S}=\mathcal{S}_1 \cup \mathcal{S}_2$, $\{ p_{1}, \ldots,p_{|\mathcal{S}|}\}$ be all the predicates in $\mathcal{S}$, we require:
\begin{eqnarray*} \label{eq:minsup}
sup(\mathcal{S}) > \max \Big( \theta, \gamma \min \big( sup(p_{1}), \ldots,sup(p_{|\mathcal{S}|}) \big) \Big)
\end{eqnarray*}where $\theta \in (0,1)$ and $\gamma \in (0,1)$ are user defined thresholds. Intuitively, the above condition means that the support of the invariant rule must be larger than a global threshold $\theta$ to achieve a minimum statistical significance. Moreover, since the support of different predicates can vary significantly in our context, we set a different minimum support threshold for each rule. More specifically, as $sup(\mathcal{S}) \leq \min(sup(p_{1}), \ldots,sup(p_{|\mathcal{S}|}) )$ according to the anti-monotone
property (the support of an itemset cannot exceed the support of its subset), we also require the support of the invariant rule to be larger than its specific upper bound scaled by $\gamma$. Regarding the $100\%$ confidence condition, we require $\frac{sup(\mathcal{S}_1\cup \mathcal{S}_2) }{sup(\mathcal{S}_1)} = 100\%$.

\section{Learning Invariant Rules}
\label{sec:learning}
Learning invariant rules from a given dataset $\mathcal{D}$ mainly consists of two steps: \emph{predicate generation} and \emph{invariant rule mining}. Specifically, we first derive the global predicate set $\mathcal{P}$ in the predicate generation step. Then, we transform each data point $\mathbf{d} \in \mathcal{D}$ as a set of satisfied predicates $S$ such that $S \subseteq \mathcal{P}$. Let $S_1,\ldots,S_{|\mathcal{D}|}$ be the records of satisfied predicates for all the data points in $\mathcal{D}$, we mine invariant rules from the records in the invariant rule mining step. Besides, we also generate univariate invariant rules for variables to detect values that are simply outside their normal range.

\subsection{Predicate Generation}
The quality of the global predicate set $\mathcal{P}$ is of vital importance for our method. Specifically, the generated predicates should have high likelihoods leading to invariant rules. Furthermore, the form of generated predicates should be as simple as possible to maximize the interpretability of invariant rules. With the above points in mind, we propose two algorithms for generating predicates, one specifically for categorical variables and the other for continuous variables. Before presenting the algorithms, it is worth noting that to ensure that any generated predicate $p$ has nonzero likelihood to contribute to an invariant rule, it is required that $sup(p) > \theta$. It can be easily seen that if the support of a predicate $p$ is less than $\theta$, then $p$ has zero likelihood to contribute to any invariant rule because any invariant rule containing $p$ cannot satisfy the minimum support condition.

\subsubsection{Predicate Generation for Categorical Variables}
The algorithm of generating predicates for categorical variables is rather straightforward. Let $\{ 1,\ldots,C \}$ be the set of seen values for a categorical variable $U_i$ in $\mathcal{D}$ and $c \in \{ 1,\ldots,C \}$, we generate a candidate predicate $p:U_i = c$. If $sup(p) > \theta$, we add $p$ to $\mathcal{P}$. Otherwise, we add $p$ to a list $\mathcal{L}$ which stores candidate predicates whose support are less than the threshold. Then, we traverse all predicates in $\mathcal{L}$ and use the ``or''($``|"$) operator to generate combined predicates until their supports are larger than the threshold. For example, let $sup(p_1)<\theta$ and $sup(p_2)<\theta$, we generate a combined predicate $p:p_1 | p_2$ if $sup(p_1 | p_2) > \theta$. Algorithm~\ref{alg:algorithmcate} gives the details of generating predicates for categorical variables. It is worth noting that by the condition checking at Line 16, the algorithm guarantees that each predicate in $\mathcal{L}$ will be included in a combined predicate as long as $sup(p_1|..|p_{|\mathcal{L}|})> \theta$.

\begin{algorithm}[tb]
  \caption{Predicate generation for categorical variables}
  \label{alg:algorithmcate}
  \begin{algorithmic}[1]
  	\REQUIRE The dataset $\mathcal{D}$, the minimum support threshold $\theta$
  	\STATE $\mathcal{P}\gets \emptyset$,$\mathcal{L}\gets \emptyset$
  	\FOR{$i=1,\ldots,N$}
  	    \STATE Let $\{1,\ldots,C\}$ be the set of possible values for $U_i$
  	    \FOR{$c=1,\ldots,C$}
          	\STATE Generate predicate $p$: $U_i=c$
          	\IF{$sup(p)>\theta$}
          	    \STATE Add $p$ to $\mathcal{P}$
          	\ELSE
          	    \STATE Add $p$ to $\mathcal{L}$
          	\ENDIF
        \ENDFOR
    \ENDFOR
    
    \STATE $k \gets 1$
    \FOR{$j=2,\ldots,|\mathcal{L}|$}
        \IF{$sup(p_k |\dots| p_j) > \theta$}
            \IF{$sup(p_{j+1} |\dots| p_{|\mathcal{L}|}) > \theta$}
                \STATE Generate predicate $p$: $p_k |\dots| p_j\quad$  
                \STATE Add $p$ to $\mathcal{P}$
                \STATE $k \gets j+1$
            \ELSE
                \STATE Generate predicate $p$: $ p_k |\dots| p_{|\mathcal{L}|}$
                \STATE Add $p$ to $\mathcal{P}$
                \STATE break
            \ENDIF
        \ENDIF
    \ENDFOR
  \RETURN $\mathcal{P}$
  \end{algorithmic}
\end{algorithm}

\subsubsection{Predicate Generation for Continuous Variables}
To maximize interpretability, we generate predicates for each continuous variable by a set of proposed cut-off values. For example, assuming there are three cut-off values $\tau_1,\tau_2,\tau_3$ for a variable $X_i$ where $\tau_1 < \tau_2 < \tau_3$, we generate four predicates which are $X_i <\tau_1$, $\tau_1 \leq X_i <\tau_2$, $\tau_2 \leq X_i <\tau_3$ and $X_i \geq \tau_3$ if the support for each predicate is larger than $\theta$. More importantly, predicates generated by such cut-off values should also lead to invariant rules with high probabilities.

To propose such cut-off values, we learn two sets of DT models using the dataset $\mathcal{D}$. Specifically, regarding the first set, we learn a DT classification model $DT(X) \rightarrow U_i$ for each categorical variable $U_i$ where all the continuous variables $X$ are used as input features and $U_i$ as the prediction target. Regarding the second set, we learn a DT regression model $DT(X_{-i}) \rightarrow X_i$ for each continuous variable $X_i$ where all the remaining continuous variables $X_{-i}$ are used as input features. Information gain and variance reduction \footnote{For DT regression models, the target variable is standardized before training such that variance reduction for different variables are comparable.} are used to measure the quality of a split for the classification models and the regression models, respectively. We train DT models with the stop criterion that the minimum number of samples required at a leaf node is larger than $|\mathcal{D}|\times \theta$. In this way, we avoid creating too many cut-off values leading to predicates with support lower than $\theta$.

Since each internal node in a learned DT model defines a split rule $\mathbf{1}_{(\tau,\infty)}(x_i)$ for a single variable $X_i$, we can extract a tuple $(X_i,\tau,q_{\tau})$ from an internal node where $\tau$ is the cut-off value for $X_i$, $q_{\tau}$ is the impurity decrease of the node. More specifically,
\begin{eqnarray*}
    q_{\tau} = \frac{N}{|\mathcal{D}|} ( H - \frac{N_{left}}{N} H_{left} - \frac{N_{right}}{N} H_{right} )
\end{eqnarray*}where $N$, $N_{left}$ and $N_{right}$ are the number of data points reaching the node, its left child and its right child; $H$, $H_{left}$ and $H_{right}$ are the impurity of the target variable for data reaching the node, its left child and right child, respectively. More specifically, impurity at a node is calculated as the entropy of the target variable for the data reaching the node for DT classification models and variance of the target variable for the data reaching the node for DT regression models. It is beneficial to select cut-off values with higher $q_{\tau}$ values to generate predicates as they reduce more uncertainty for the corresponding target variables thus are more likely to contribute to invariant rules. 

\begin{figure}[tb]
\centering
\includegraphics[width=.45\textwidth]{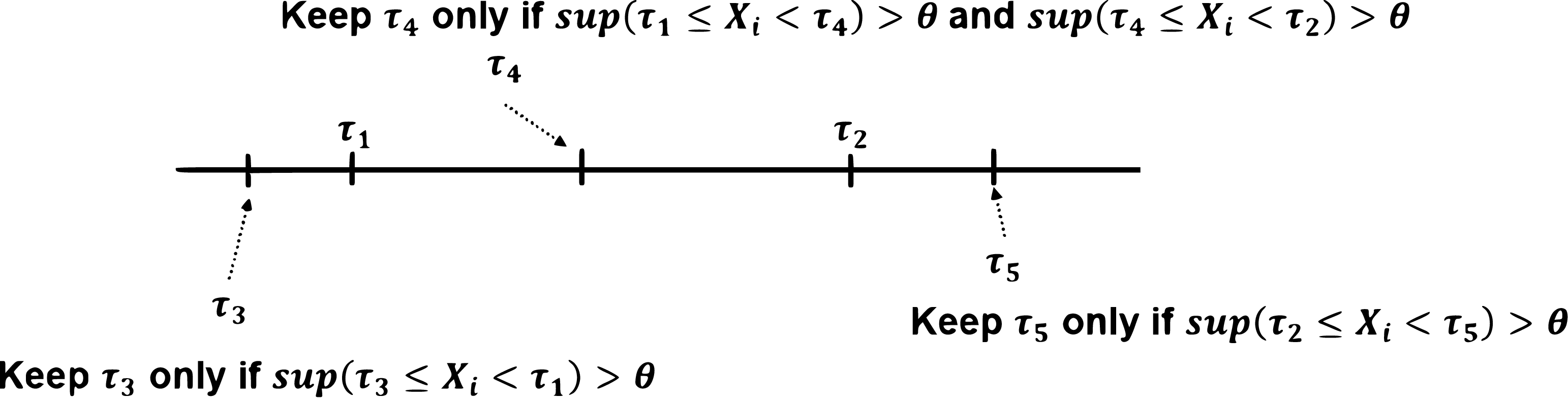}
\caption{Illustration for the logic of whether include a cut-off value with a lower $q_{\tau}$ value for predicate generation.}
\label{fig:cut-off}
\end{figure}

Therefore, after extracting all the cut-off values for a continuous variable $X_i$ from all the trained DT models, we arrange $X_i:(\tau_1, \dots,\tau_J)$ for the continuous variable such that $q_{\tau_1} \geq...\geq q_{\tau_J}$, i.e., the $q_{\tau}$ value for the cut-off values are sorted in a descending order. Then, we sequentially traverse the list of cut-off values and only keep a cut-off value for predicate generation if its inclusion will not cancel the predicate of a cut-off value with a higher $q_{\tau}$ value. The logic of whether include a cut-off value for predicate generation is illustrated in Figure~\ref{fig:cut-off}. Algorithm~\ref{alg:predicate_cont} gives the details of predicate generation for continuous variables.

\begin{algorithm}[tb]
  \caption{Predicate generation for continuous variables}
  \begin{algorithmic}[1]
  	\REQUIRE The dataset $\mathcal{D}$, the global predicate set $\mathcal{P}$, the minimum support threshold $\theta$
  	\STATE $\mathcal{T}\gets \emptyset$
  	\FOR{$i=1,\ldots,N$}
        \STATE Train a DT classification model $DT(X) \rightarrow U_i$
        \STATE For each internal node $j$ in the trained DT model, add $(X_j,\tau_j, q_{\tau_j})$ to $\mathcal{T}$ based on its split rule
    \ENDFOR
    
    \FOR{$i=1,\ldots,M$}
        \STATE Train a DT regression model $DT(X_{-i}) \rightarrow X_i$
        \STATE For each internal node $j$ in the trained DT model, add $(X_j,\tau_j, q_{\tau_j})$ to $\mathcal{T}$ based on its split rule 
    \ENDFOR
    
    \FOR{$i=1,\ldots,M$}
        \STATE Get $X_i:(\tau_1, \dots,\tau_J)$ from $\mathcal{T}$ where $q_{\tau_1} \geq ...\geq q_{\tau_J}$
        \STATE $\mathcal{L}  \gets [ \tau_1 ]$ \COMMENT{list of cut-off values to keep}
        \FOR{$j=2,\ldots,J$}
            \STATE $k \gets 1$ \COMMENT{insert position of $\tau_j$}
            \WHILE{$k \leq |\mathcal{L}|$ and $\tau_j < \mathcal{L}[k]$ }
                \STATE $k \gets k+1$
            \ENDWHILE
            
            \IF{$k=1$}
                 \IF{$sup( \tau_j \leq X_i < \mathcal{L}[1]) > \theta$}
                    \STATE Insert $\tau_j$ to $\mathcal{L}$ at position $k$
                \ENDIF
            \ELSIF{$k=|\mathcal{L}|+1$}
                \IF{$sup( \mathcal{L}[k-1]\leq X_i < \tau_j ) > \theta$}
                    \STATE Insert $\tau_j$ to $\mathcal{L}$ at position $k$
                \ENDIF
            \ELSE
                \IF{$sup( \mathcal{L}[k-1]\leq X_i < \tau_j ) > \theta$ and $sup( \tau_j \leq X_i < \mathcal{L}[k]) > \theta$}
                    \STATE Insert $\tau_j$ to $\mathcal{L}$ at position $k$
                \ENDIF
            \ENDIF
        \ENDFOR
        \FOR{$j=1,\ldots,|\mathcal{L}|$}        
            \IF{$j=1$} 
                \STATE Generate predicate $p: X_i < \mathcal{L}[j]$ and add $p$ to $\mathcal{P}$
            \ENDIF
            \IF{$ 1<j \leq |\mathcal{L}|$}
                \STATE Generate predicate $p: \mathcal{L}[j-1] \leq X_i < \mathcal{L}[j]$ and add $p$ to $\mathcal{P}$
            \ENDIF
            \IF{$j=|\mathcal{L}|$}
                \STATE Generate predicate $p: X_i \geq \mathcal{L}[j]$ and add $p$ to $\mathcal{P}$
            \ENDIF
        \ENDFOR
    \ENDFOR
\RETURN $\mathcal{P}$
  \end{algorithmic}
   \label{alg:predicate_cont}
\end{algorithm}

\subsection{Invariant Rule Mining}
With the global predicate set $\mathcal{P}$, we transform each data point $\mathbf{d} \in \mathcal{D}$ as a set of satisfied predicates $S$ such that $S \subseteq \mathcal{P}$. Given the records of satisfied predicates for the data in $\mathcal{D}$ as $S_1,\ldots,S_{|\mathcal{D}|}$, mining invariant rules is a well-studied problem in association rule mining. Concretely, we first find all frequent predicate sets with multiple minimum support thresholds from the records using algorithms such as CFP-growth \cite{hu2006mining}. Then, for an arbitrary frequent predicate set $\mathcal{S}$, we randomly partition it into two non-empty sets $\overline{\mathcal{S}}$ and $\mathcal{S}-\overline{\mathcal{S}}$, an invariant rule $\overline{\mathcal{S}} \Rightarrow \mathcal{S} - \overline{\mathcal{S}}$ is generated if its confidence is $100\%$, i.e., $sup(\mathcal{S})/sup(\overline{\mathcal{S}}) = 100\%$.

\subsection{Univariate Invariant Rule Generation}
We also generate univariate invariant rules for variables to detect values that are outside their normal range. Concretely, for each categorical variable $U_i$, we generate an invariant rule $\emptyset \Rightarrow U_i \in \{1,\ldots,C\}$, where $\{1,\ldots,C\}$ is the set of seen values for $U_i$ in $\mathcal{D}$. For each continuous variable $X_i$, we generate an invariant rule $\emptyset \Rightarrow \min (\mu_i-3\sigma_i, \underline{x_i}) \leq X_i \leq \max (\mu_i+3\sigma_i,\overline{x_i})$, where $\mu_i$ and $\sigma_i$ are the mean and standard deviation of $X_i$ in $\mathcal{D}$, $\underline{x_i}$ and $\overline{x_i}$ are the smallest value and largest value of $X_i$ in $\mathcal{D}$.

\section{Anomaly Detection and Interpretation}
\label{sec:inference}
In this section we introduce how to utilize learned invariant rules for anomaly detection and interpretation in the inference phase. Specifically, we first introduce how to calculate an anomaly score for a data point using learned invariant rules. Then, we introduce what useful information is conveyed by the violated invariant rules regarding reported anomalies.


\subsection{Anomaly Score Calculation}
During the inference phase, we say an invariant rule is violated by a data point if all the predicates in the antecedent set are satisfied by the data point and at least one predicate in the consequent set is not satisfied. Once an invariant rule is violated by a data point, we increase some amount of anomaly score for the data point and the amount of anomaly score is calculated by multiplying the support of the violated rule and the degree of violation to the rule. Specifically, let $l:\mathcal{S}_1 \Rightarrow \mathcal{S}_2$ be an invariant rule, $sup(l)$ be the support of the rule, $\mathcal{S}_2=\{ p_1,..., p_{|\mathcal{S}_2|}\}$, $\mathbf{d}$ be a data point which violates $l$, then the anomaly score induced by $l$ to $\mathbf{d}$, denoted as $f(l,\mathbf{d})$, is calculated as follows:
\begin{equation*}
    f(l,\mathbf{d}) = sup(l) \times \sum_{i=1}^{|\mathcal{S}_2|} g(p_i,\mathbf{d}) 
\end{equation*}where
\[ g(p,\mathbf{d})  =
  \begin{cases}
    1       & \quad \text{if } p: U_i = c \text{ and } u_i \neq c\\
    1       & \quad \text{if } p: p_1 | ...| p_j \text{ and $p_1,...,p_j$ not satisfied by $\mathbf{d}$}  \\
    1       & \quad \text{if } p: U_i \in \{1,\ldots,C\} \text{ and } u_i \notin \{1,\ldots,C\}\\
    \frac{x_i-\tau}{\sigma_i}       & \quad \text{if } p: X_i < \tau \text{ and } x_i > \tau \\
    \frac{\tau-x_i}{\sigma_i}       & \quad \text{if } p: X_i \geq \tau \text{ and } x_i < \tau \\
    \frac{x_i-\tau_2}{\sigma_i}       & \quad \text{if } p: \tau_1 \leq X_i < \tau_2 \text{ and } x_i > \tau_2 \\
    \frac{\tau_1-x_i}{\sigma_i}       & \quad \text{if } p: \tau_1 \leq X_i < \tau_2 \text{ and } x_i < \tau_1 \\
    0      & \quad \text{otherwise } \\
  \end{cases}
\]
In the above equation, $g(p,\mathbf{d})$ denotes the degree of violation of data point $\mathbf{d}$ to the predicate $p$, and it is 1 if a predicate of categorical variable is violated; the distance between the bound and the observed value over the standard deviation of the variable in $\mathcal{D}$ if a predicate of continuous variable is violated. For example, this means $x_i=4$ will get a larger anomaly score than $x_i=3.5$ by violating the same predicate $X_i<3$. Furthermore, The degree of violation of data point $\mathbf{d}$ to the whole rule is calculated by summing up the degree of violation of the data point to all the predicates in the consequent set of the rule. Intuitively, we impose a larger anomaly score to the data point if it breaks a rule with a larger support and if it violates the rule with a larger extent. We gives the details of anomaly score calculation for a data point in Algorithm~\ref{alg:algorithm_score}. 

Once an anomaly score $s$ has been calculated for a data point $\mathbf{d}$, we report $\mathbf{d}$ as an anomaly if $s > \varphi$ where $\varphi$ is a user-defined threshold. Particularly, setting $\varphi=0$ means that we will report $\mathbf{d}$ as an anomaly if it breaks any invariant rule. In practice, an anomaly threshold can be tuned by methods such as setting an acceptable false positive rate on a validation dataset.

\begin{algorithm}[tb]
  \caption{Anomaly score calculation}
  \label{alg:algorithm_score}
  \begin{algorithmic}[1]
  	\REQUIRE The learned invariant rules $\mathcal{L}$, a data point $\mathbf{d}$
  	\STATE Initialize anomaly score $s \leftarrow 0$
  	\STATE Initialize the violated invariant rules $\mathcal{V} \leftarrow \emptyset$
  	\FOR{$l$ in $\mathcal{L}$}
  	    \IF{$l$ is violated by $\mathbf{d}$}
      	    \STATE $s \leftarrow s+f(l,\mathbf{d})$
      	    \STATE Add $l$ to $\mathcal{V}$
      	\ENDIF
    \ENDFOR
  \RETURN $s$, $\mathcal{V}$
  \end{algorithmic}
\end{algorithm}

\subsection{Anomaly Interpretation}
Importantly, as can be seen from Algorithm~\ref{alg:algorithm_score}, apart from the anomaly score for the data point, our algorithm also reports the violated invariant rules of the data point. This gives the chance of further anomaly interpretation. Specifically, as the invariant rules are self-explainable, they can provide extremely useful information regarding the reported anomaly. For example, assume the invariant rule $l: 5\leq X_1 <10, X_2>20.4  \Rightarrow  X_3<7.1 $ is violated by the reported anomaly, the following three critical questions can be answered to interpret the reported anomaly:
\begin{itemize}
\item Q1: What specific feature(s) is abnormal:
\item A1: It is mostly likely to be $X_3$ according to $l$.
\item Q2: What normal value should be for the abnormal feature?
\item A2: $X_3$ should be less than $7.1$ according to $l$.
\item Q3: Why the algorithm thinks $X_3$ should be less than $7.1$?
\item A3: Because we find in the training data $\mathcal{D}$, $5\leq X_1 <10$, $X_2>20.4$ and $X_3<7.1$ are simultaneously satisfied many times (larger than the minimum support threshold). Moreover, when $5\leq X_1 <10$ and $X_2>20.4$ occur, then $X_3<7.1$ must co-occur (according to the $100\%$ confidence condition).
\end{itemize}To our knowledge, there is no such data-driven anomaly detection method that can answer the above three critical questions in such an interpretable way as ours. In practice, the straightforward clues provided by the violated invariant rule(s) can be very useful for conducting root cause analysis for the reported anomaly.

\section{Experiments}
\label{sec:experiments}
This section presents the detailed results for three experiments designed to evaluate the effectiveness of our proposed method. In the first experiment, we compare the anomaly detection performance of our method with some state-of-the-art anomaly detection models on various public benchmark datasets. We also demonstrate the importance of our DT-based predicate generation method by comparing its performance with other simplified predicate generation methods including KMeans and uniform interval-based. In the second experiment, we study the impact of key hyperparameters in our method. In the third experiment, we study whether the learned invariant rules can provide accurate clues for interpreting the reported anomalies.

Before giving experimental details, we first give a brief time efficiency analysis of our method. The training process of our method mainly consists of two steps: predicate generation and invariant rule mining. Since the predicate generation step has almost linear time complexity, the time cost during the training phase is dominated by the invariant rule mining step or the algorithm for mining frequent predicate sets with multiple minimum support thresholds to be more specific. The time complexity of such mining algorithms is largely impacted by the characteristics of the dataset, thus does not have a formal big $O$ style formulation. In our experiment, we choose CFP-growth as the mining algorithm due to its implementation simplicity and proved efficiency. We refer to~\cite{hu2006mining} for a more detailed efficiency analysis of the CFP-growth algorithm. In the inference phase, one can easily find that the time complexity is linear with the number of learned invariant rules, and the process can be easily paralleled.

\subsection{Performance Analysis}

\begin{table*}[tb]
\begin{center}
\begin{tabular}{lcccccc}
\hline
& Domain &Train size & Test size &  Num. cont. vars. & Num. cat. vars & Anomalies($\%$) \\
\hline
SWAT& Industrial process monitoring & 99360 & 89984 &  50 & 15 & 12 \\
BATADAL& Industrial process monitoring & 8761 & 4177 &  27 & 16 & 5 \\
KDDCup99& Network intrusion detection  &78416 & 23577 &  32 & 6 & 20 \\
Gas pipeline& Network intrusion detection  & 129037 & 106925 &  9 & 8 & 20 \\
Annthyroid& Disease detection & 3998 & 2880 &  6 & 15 & 7 \\
Cardio&  Disease detection & 1099 & 696 &  19 & 2 & 20 \\
\hline
\end{tabular}
\caption{The details of benchmark datasets.}
\label{tab:datasets}
\end{center}
\end{table*}

\begin{table*}[tb]
\begin{center}
\begin{tabular}{lcc|cc|cc|cc|cc|cc|cc}
\hline
\multirow{2}{*}{} &  \multicolumn{2}{c}{LOF}  &  \multicolumn{2}{c}{IF} &  \multicolumn{2}{c}{AE} &  \multicolumn{2}{c}{DeepSVDD} & \multicolumn{2}{c}{IR UniformBins} & \multicolumn{2}{c}{IR KMeansBins}& \multicolumn{2}{c}{Ours}\\
& AUC & pAUC  & AUC & pAUC & AUC & pAUC  & AUC & pAUC & AUC & pAUC  & AUC & pAUC & AUC & pAUC  \\
\hline
SWAT &  0.69 & 0.49  & \textbf{0.85} & 0.82 &  0.84 & \textbf{0.83} &0.50& 0.50 & 0.79 & 0.57& 0.78 & 0.68& 0.81 & 0.81 \\
BATADAL &  \textbf{0.81} & \textbf{0.72}&  0.71& 0.57 &  0.66& 0.58 & 0.67 & 0.67 & 0.65 & 0.52 & 0.70 & 0.57 & 0.79 & \textbf{0.72} \\
KDDCup99 & 0.93& 0.63 & 0.95& 0.74 & 0.99 & 0.96 & 0.99 & 0.98 & 0.81& 0.81 & 0.81 & 0.80& \textbf{1.00} & \textbf{0.99}\\
Gas pipeline & 0.65 & 0.53  & \textbf{0.71}& 0.57 & 0.70& 0.63 & 0.66& 0.62& 0.64& 0.64 & 0.64 & 0.64& 0.69 & \textbf{0.69}\\
Annthyroid & 0.73& 0.54 &  0.62& 0.50 & \textbf{0.77} & \textbf{0.62} & 0.50&0.52 & 0.57& 0.57 & 0.58 & 0.58& 0.61 & 0.59\\
Cardio  & 0.92& 0.75 &  0.94& 0.75 & 0.94 & 0.80 &0.79&0.61& 0.92& 0.80& 0.93 & 0.79& \textbf{0.95} & \textbf{0.88}\\
\hline
Avg.  & 0.79 & 0.61 &  0.80 & 0.66 & \textbf{0.82} & 0.74 &0.69 & 0.65 & 0.73 & 0.65 & 0.74 & 0.68 & 0.81 & \textbf{0.78}\\
Avg. Rank & 4.00 & 5.33 & \textbf{2.33} & 5.17 & 2.50 & 2.50 & 5.33 & 4.83 & 5.67 & 4.00 &5.00 & 3.83 & \textbf{2.33} & \textbf{1.50} \\
\hline
\end{tabular}
\caption{Results, in terms of AUC and standardized pAUC, for the anomaly detection models on the benchmark datasets.}
\label{tab:auc_comp}
\end{center}
\end{table*}

\subsubsection{Baseline models}
We compared the performance of our proposed method with four state-of-the-art anomaly detection models, namely: Local Outlier Factor (LOF) \cite{breunig2000lof}, Isolation Forest (IF) \cite{liu2012isolation}, Autoencoder (AE) and Deep Support Vector Data Description (DeepSVDD) \cite{ruff2018deep}. Specifically, LOF calculates anomaly scores by measuring the local deviation of density of a given sample with respect to its neighbors. IF calculates anomaly scores by measuring how easy a sample can be isolated with others using tree-based ensembles. AE is the vanilla version of reconstruction-based method based on neural networks where the errors of reconstructed samples are used as anomaly scores and it often plays as the backbone of more advanced deep anomaly detection models. DeepSVDD trains a neural network which aims to squeeze all training samples into a hypersphere whose radius is as small as possible and the radius of samples are used as anomaly scores in the inference phase. Furthermore, we replace our DT-based predicate generation method for continuous variables with an uniform interval-based and a KMeans-based discretization method which discretize continuous variables to 5 bins for predicatie generation. To demonstrate the importance of our DT-based predicate generation method, we also compare the performance of our original method with the two simplified methods which we call invariant rules with uniform bins (IR UniformBins) and invariant rules with KMeans bins (IR KMeansBins).

We implement LOF and IF using Scikit-Learn \cite{pedregosa2011scikit} library of version 1.1.2. The default hyperparameter values in Scikit-Learn are used. For AE, the number of neurons in the bottleneck layer is set to $\frac{1}{4}$ of the input dimension, and other hyperparameters such as number of hidden layers, number of training epochs are tuned to minimize the reconstruction error using cross validation. For DeepSVDD, we use a two-hidden-layer neural network and the number of neurons in the first and second hidden layers is set to $\frac{1}{2}$ and $\frac{1}{4}$ of the input dimension respectively, the $\nu$ hyperparameter is set to 0.01 and the soft-boundary Deep SVDD objective \cite{ruff2018deep} is used. Regarding IR UniformBins, IR KMeansBins and our method, we firstly split out $20\%$ of training data as a validation dataset. We explicitly set $\gamma$ to 0.7 for all datasets, and $\theta$ is tuned to the smallest value with which the false positive rate is lower than 0.01 on the validation dataset when setting $\varphi=0$, i.e., report a data point as an anomaly if it breaks any learned invariant rule. Moreover, we set the maximum number of predicates in a rule to 5 such that the learned invariant rules are not too complicated to understand.


\subsubsection{Benchmark datasets}
We select six benchmark datasets in three application domains: SWAT \cite{goh2016dataset} and BATADAL \cite{taormina2018battle} for industrial process condition monitoring, KDDCup99~\cite{kdd99} and Gas Pipeline \cite{morris2015industrial} for network intrusion detection, Annthyroid \cite{Rayana2016odds} and Cardio \cite{Rayana2016odds} for disease detection. For SWAT, we create a new feature $\Delta X_i = X_i (t+1) - X_i (t)$ for each continuous variable to account for time series update of sensor values. For KDDCup99, the 10 percent version is used. For all datasets which have anomaly ratio that is larger than $20\%$, we randomly drop anomalies until the anomaly ratio reaches $20\%$. Moreover, all the training set in our experiments contains no anomalies. More details of the benchmark datasets are given in Table~\ref{tab:datasets}.

\subsubsection{Performance metrics and results} The AUC ROC is commonly used as the metric to compare the performance of different anomaly detection algorithms. However, one of the major practical drawbacks of the AUC is that it summarizes the entire ROC curve, including regions with a false positive rate that is far too high for practical applications. As a result, we also report the standardized partial AUC (pAUC) \cite{mcclish1989analyzing} with a false positive rate that is lower than 0.1 to compare the performance of algorithms with a low false positive rate. 

The experiment results for the anomaly detection models on the benchmark datasets are reported in Table~\ref{tab:auc_comp}. Our method achieves best AUC on 2 datasets, best pAUC on 4 datasets. To summarize, our method achieves similar average performance in terms of AUC compared with best performing baselines IF and AE, and outperforms baseline models with a clear margin in terms of average pAUC. This observation indicates that using invariant rules can effectively detect anomalies with a low false positive rate which is a significant advantage in practical applications. Furthermore, our method also outperforms IR UniformBins and IR KMeansBins with a clear margin. This demonstrates the importance of our DT-based predicate generation method for continuous variables.


\subsection{Impact of Hyperparameters}

In the second experiment, we examine the impact of different values of $\theta$ and $\gamma$ on the anomaly detection performance of our method. Specifically, we experiment with $\theta=[0.01,0.02,0.04,0.08,0.16,0.32]$, $\gamma=[0,0.3,0.6,0.9]$ on all the benchmark datasets. 

\begin{figure*}[tb]
\centering
\includegraphics[width=.32\textwidth]{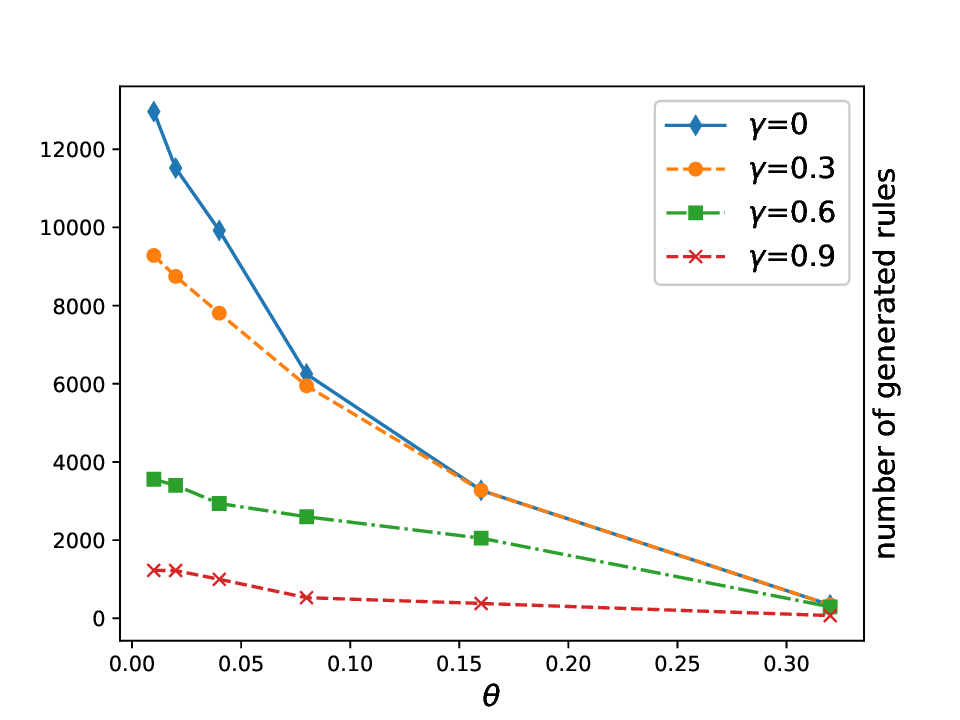}%
\hfil
\includegraphics[width=.32\textwidth]{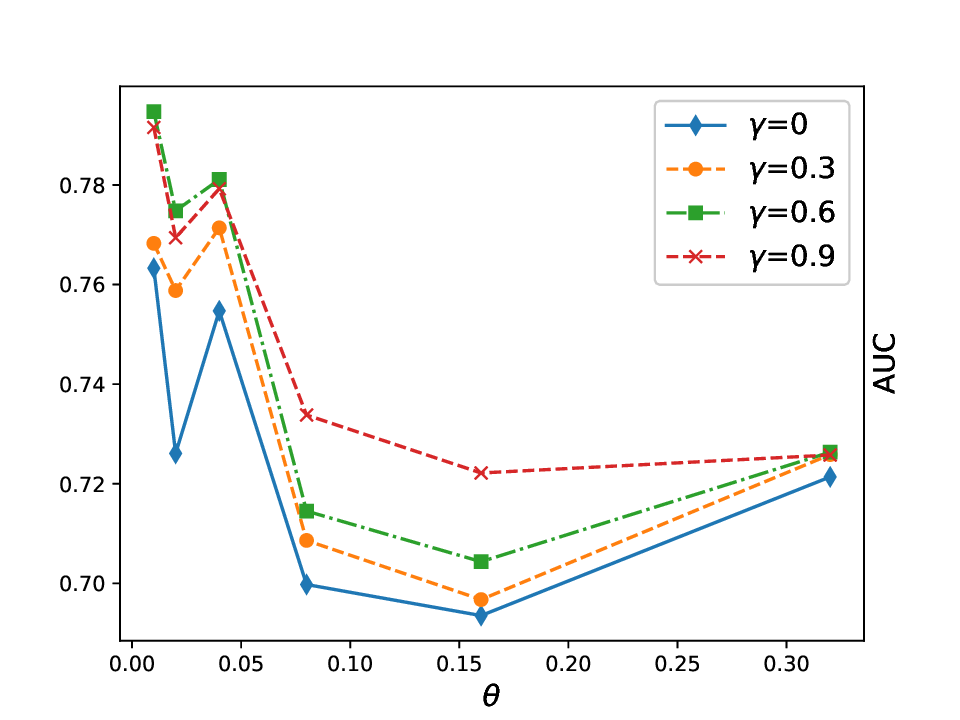}
\hfil
\includegraphics[width=.32\textwidth]{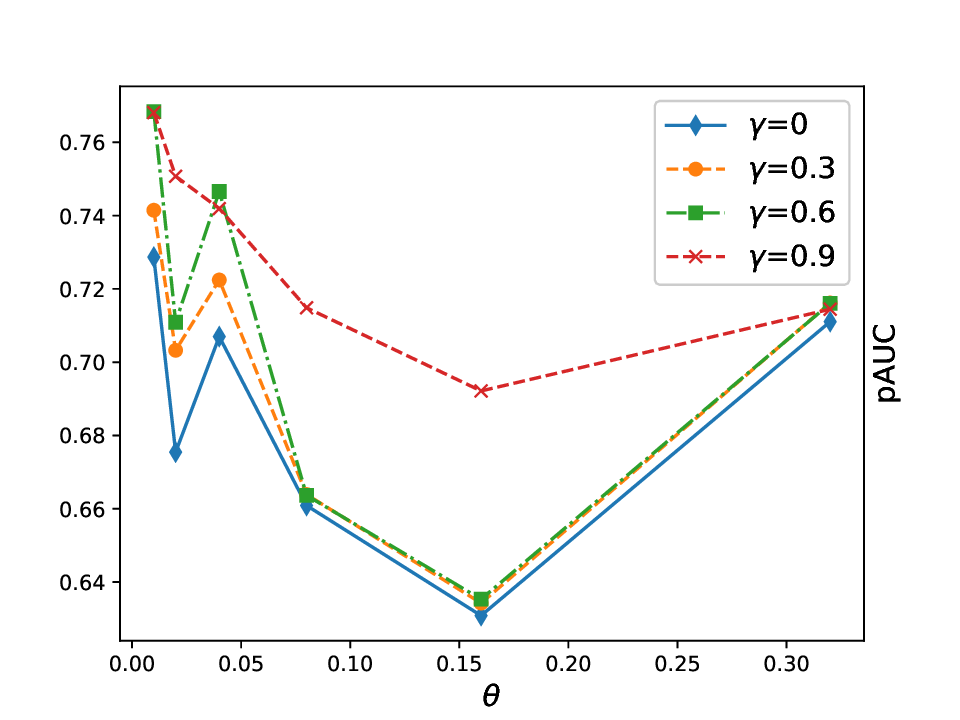}
\caption{(left): the average number of generated invariant rules with different values of $\theta$ and $\gamma$ for the benchmark datasets; (right): the average AUC with different values of $\theta$ and $\gamma$ for the benchmark datasets; (right): the average pAUC with different values of $\theta$ and $\gamma$ for the benchmark datasets;}
\label{fig:hpo}
\end{figure*}

It is conceivable that more invariant rules will be learned with smaller values for $\theta$ and $\gamma$, thus more anomalies can be potentially detected. However, the learned invariant rules will also have lower statistical significance which will lead to a higher false positive rate. Therefore, setting proper values for $\theta$ and $\gamma$ are of vital importance in our method. In Figure~\ref{fig:hpo}, we plot the average number of generated invariant rules, avarge AUC and pAUC on the benchmark datasets with different values of $\theta$ and $\gamma$. As can be seen from the figures, our method generally achieves better performance with a smaller value for $\theta$ and a larger value for $\gamma$. This is because setting a small value for $\theta$ allows to generate more invariant rules which is beneficial for detecting anomalies in general. Additionally, it is also rather important to control the ``quality" of learned rules by setting a large value for $\gamma$. Specifically, consider that there is a predicate $p_1$ with a small support but still considerably larger than $\theta$, another predicate $p_2$ with a very large support. In this case, it is very easy to generate a rule $p_1 \Rightarrow p_2$ just because of coincidence if $\gamma$ takes a small value. Such rules can be effectively filtered out by setting a large value to $\gamma$ because we require $sup(\{ p_1,p_2 \}) > \gamma\times sup(p_1)$, i.e., at least $\gamma$ proportion of the data points satisfying $p_1$ must also satisfy the rule. To sum up, we recommend to set a large value, e.g., 0.7, for $\gamma$ in general. Furthermore, we recommend to set the value for $\theta$ to the smallest value with which the false positive rate is lower than an acceptable threshold on the validation dataset when setting $\varphi=0$.

\begin{table*}[tb]
\begin{center}\begin{footnotesize}
\begin{tabular}{llc}
\hline
Attack segment& Target sensors/actuators &  Violated invariant rules \\
\hline
1 & MV101 & $ P203=2, 1.07<=FIT101<2.48  \Rightarrow  MV201=2,\color{red}MV101=2\color{black} $ \\
\hline
2 & P101,P102 & $884.79<=LIT401<910.69, P205=2, 804.58<=LIT301<934.12 \Rightarrow  MV101=2, \color{red}P101=2\color{black}$ \\
\hline
3 & LIT101,P101 & $DPIT301<7.50 , \color{red}P101=1\color{black} \Rightarrow FIT301<1.94 , MV302=1$ \\
\hline
4 & MV504 & --\\
\hline
\multirow{4}{*}{5} & \multirow{4}{*}{AIT202,P203} & $P205=2 \Rightarrow \color{red}P203=2 \color{black}$\\
&   & $\emptyset \Rightarrow \color{red}8.12<=AIT202<=8.99 \color{black}$ \\
 &   & $\emptyset \Rightarrow \color{red}-0.12<=\Delta AIT202<=0.16 \color{black} $ \\
 &   & $\color{red}P203=1 \color{black} \Rightarrow P205=1 $\\
\hline
\multirow{2}{*}{6} & \multirow{2}{*}{LIT301,P301} & $ \color{red}LIT301>=1004.59 \color{black}, MV101=2  \Rightarrow P101=1, MV302=1  $\\
&   & $ MV302=2 , AIT202<8.40 , AIT401>=148.81 \Rightarrow MV201=2 , P101=2$ \\
 \hline
 \multirow{4}{*}{7} & \multirow{4}{*}{DPIT301,LIT301,LIT401} & $\emptyset \Rightarrow  \color{red}-3.73<=DPIT301<=36.86 $\\
&   & $\emptyset \Rightarrow  \color{red}-13.39<=\Delta DPIT301<=6.01\color{black}$ \\
&  & $ \color{red}DPIT301>=20.11 \color{black}\Rightarrow MV302=2$ \\
&  & $ \color{red}LIT301>=1004.59 \color{black}, \color{red}-0.08<=\Delta DPIT301<0.09 \color{black}, FIT301<1.94 \Rightarrow P302=1 , P101=1$  \\
 \hline
\multirow{2}{*}{8} & \multirow{2}{*}{FIT401,P501} & $\emptyset \Rightarrow 132.59<=AIT402<=235.71 $\\
&   & $\emptyset \Rightarrow \color{red}-0.59<=\Delta FIT401<=0.84 \color{black}$  \\
\hline
9  & MV304 & --\\
 \hline
\multirow{2}{*}{10} & \multirow{2}{*}{MV303} & $ 329.26<=AIT203<332.17 , MV101=2 , AIT402<170.28 \Rightarrow AIT502<166.59 , AIT202<8.40 $\\
&  & $8.40<=AIT202<8.44, \Delta AIT202<0.0018 , P101=1 \Rightarrow FIT201<2.00, P205=1 $  \\
\hline
11 & LIT301  & --\\
\hline
\multirow{2}{*}{12} & \multirow{2}{*}{MV303} & $ 329.26<=AIT203<332.17 , MV101=2 , AIT402<170.28 \Rightarrow AIT502<166.59 , AIT202<8.40 $\\
&  & $8.40<=AIT202<8.44 , \Delta AIT202<0.0018 , P101=1\Rightarrow FIT201<2.00 , P205=1 $  \\
\hline
13 & AIT504 & --\\
 \hline
14 & AIT504 & $\emptyset \Rightarrow \color{red}-31.40 <= \Delta AIT504<=105.17 \color{black}$\\
 \hline
15  & MV101, LIT101 & $329.26<=AIT203<332.17, \color{red}MV101=2\color{black}, AIT402<170.28 \Rightarrow AIT502<166.59, AIT202<8.40$\\
 \hline
\multirow{4}{*}{16} & \multirow{4}{*}{UV401, AIT502, P501} & $\emptyset \Rightarrow \color{red}-1.67<=\Delta AIT502<=1.15 \color{black}$\\
&   & $\emptyset \Rightarrow 132.59<=AIT402<=235.71$ \\
&  & $\emptyset \Rightarrow \color{red}119.08<=AIT502<=218.33 \color{black}$ \\
&  & $\emptyset \Rightarrow 7.30=AIT501<=8.02 $\\
\hline
\multirow{4}{*}{17} & \multirow{4}{*}{P602, DPIT301, MV302} & $\color{red}-0.08<=\Delta DPIT301<0.09\color{black}, FIT301<1.94 \Rightarrow \color{red}DPIT301<7.50 \color{black}, \color{red}MV302=1 \color{black}$\\
&  & $\emptyset \Rightarrow \color{red}-3.73<=DPIT301<=36.86\color{black}$ \\
&  & $\emptyset \Rightarrow \color{red}-13.39<=\Delta DPIT301<=6.01\color{black}$ \\
&  & $\emptyset \Rightarrow 244.33<=AIT201<=278.25$\\
\hline
18 & P203, P205 &  $FIT201>=2.45 , MV302=2 \Rightarrow MV201=2 , \color{red}P203=2 \color{black}$  \\
\hline
19 & LIT401, P401 & --\\
 \hline
 \multirow{3}{*}{20} & \multirow{3}{*}{P101, LIT301,P102} & $329.26<=AIT203<332.17 , MV101=2 , AIT402<170.28 \Rightarrow AIT502<166.59 , AIT202<8.40$ \\
& & $\color{red}LIT301>=1004.59 \color{black}, MV101=2 \Rightarrow \color{red}P101=1 \color{black}, MV302=1$ \\
&  & $\color{red}LIT301>=1004.59\color{black} \Rightarrow FIT201<2.00 $\\
\hline
\multirow{4}{*}{21} & \multirow{4}{*}{P302, LIT401} & $\emptyset \Rightarrow 132.59<=AIT402<=235.71  $\\
& & $\emptyset \Rightarrow 119.08<=AIT502<=218.33$ \\
&  & $\emptyset \Rightarrow 0.001<=FIT501<=2.12$ \\
& & $\Delta LIT301<1.12 , 332.17<=AIT203<342.99 , \color{red}\Delta LIT401>=-0.96\color{black} \Rightarrow P203=2, P205=2 $\\
 \hline
22 & P201, P203, P205& -- \\
 \hline
\multirow{3}{*}{23} & \multirow{3}{*}{LIT101, P101, MV201} & $884.79<=LIT401<910.69 ,P205=2, 804.58<=LIT301<934.12 \Rightarrow MV101=2 , \color{red}P101=2\color{black} $\\
&  & $\Delta LIT301<1.12, P205=2 \Rightarrow P203=2 , \color{red}P101=2\color{black}$ \\
&  & $\color{red}MV201=2\color{black}, MV302=2, 934.12<=LIT301<964.00 \Rightarrow P203=2, MV101=2$ \\
\hline
24 & LIT-401 & --\\
 \hline
\multirow{3}{*}{25}  & \multirow{3}{*}{LIT301, P302} & $\color{red}LIT301>=1004.59\color{black} , MV101=2 \Rightarrow P101=1 , MV302=1 $\\
&  & $\color{red}LIT301>=1004.59\color{black} \Rightarrow FIT201<2.00$ \\
&  & $FIT201>=2.45 , MV302=2 \Rightarrow MV201=2 , P203=2$ \\
\hline
26 & LIT101, P101 & $884.79<=LIT401<910.69 , P205=2, 804.58<=LIT301<934.12 \Rightarrow MV101=2, \color{red}P101=2\color{black}$\\
\hline
27 & P101 & $\Delta LIT301<1.12, P205=2 \Rightarrow P203=2 , \color{red}P101=2\color{black}$\\
 \hline
\multirow{2}{*}{28} & \multirow{2}{*}{P01, P102} & $P205=2 , 804.58<=LIT301<934.12 \Rightarrow P203=2 , \color{red}P101=2 \color{black}$\\
& & $498.08<=LIT101<511.56, -0.06<=\Delta AIT203<0.12 \Rightarrow MV201=2, P203=2$ \\
\hline
29 & LIT101 &--\\
 \hline
\multirow{2}{*}{30} & \multirow{2}{*}{FIT502, P501} & $\emptyset \Rightarrow -0.004<=\Delta FIT504<=0.103 $\\
& & $\emptyset \Rightarrow -0.20<=\Delta FIT503<=0.24$ \\
 \hline
\multirow{4}{*}{31} & \multirow{4}{*}{AIT402, AIT502} & $\emptyset \Rightarrow  \color{red}-1.67<=\Delta AIT502<=1.15 \color{black}$\\
& & $\emptyset \Rightarrow  \color{red}-2.155<=\Delta AIT402<=2.23\color{black}$ \\
& & $\emptyset \Rightarrow \color{red} 119.08<=AIT502<=218.33\color{black}$ \\
&  & $\emptyset \Rightarrow  \color{red}132.59<=AIT402<=235.71\color{black}$ \\
\hline
\multirow{3}{*}{32} & \multirow{3}{*}{FIT-401, AIT-502} & $\emptyset \Rightarrow \color{red}-0.59<=\Delta FIT401<=0.84 \color{black}$\\
& & $\emptyset \Rightarrow 251.20<=AIT503<=283.36$ \\
& & $\emptyset \Rightarrow \color{red}-1.67<=\Delta AIT502<=1.15\color{black}$ \\
\hline
\multirow{4}{*}{33} & \multirow{4}{*}{FIT-401} & $\emptyset \Rightarrow -0.004<=\Delta FIT504<=0.103 $\\
&  & $\emptyset \Rightarrow AIT501>=7.90 \Rightarrow AIT402<170.28$ \\
&  & $\emptyset \Rightarrow \color{red}-0.59<=\Delta FIT401<=0.84\color{black}$ \\
&  & $\emptyset \Rightarrow -0.006<=\Delta AIT501<=0.013$ \\
\hline
34 & LIT301 &--\\
\hline
\end{tabular}
\caption{The violated invariant rules during attack segments in the test set of SWAT.}
\label{tab:attack}
\end{footnotesize}\end{center}
\end{table*}
\subsection{Interpretability Analysis}
In the third experiment, we study whether the learned invariant rules can provide useful information to assist users to locate and understand the cause of the detected anomalies which is of vital importance in practice. 

Specifically, we conduct the experiment on SWAT dataset in which the anomalous data consists of 34 attack segments with clearly labeled target features (sensors and/or actuators). In total, 2484 invariant rules are learned from the training set with $\theta=0.01$ and $\gamma=0.7$. Then, we report the violated invariant rules during all the attack segments in the test set in Table~\ref{tab:attack}. As can be seen from the table, at least one invariant rule is violated in 25 out of 34 attack segments. In addition, we find at most 4 invariant rules are violated in each segment, which means that the user only needs to check at most 4 rules to interpret each reported anomaly in this case. Furthermore, we can also find that at least one violated invariant rule directly involves the target feature(s) in 22 out of those 25 detectable attack segments. Importantly, as these invariant rules are highly understandable even for a user without a data science background, they can provide extremely useful information to help users understand the cause of reported anomalies in practice.

 \section{Related Work}
 \label{sec:related}
In most anomaly detection applications, accuracy alone is not sufficient, interpretability and trustworthiness are equally critical \cite{ruff_unifying_2021}. However, building interpretable anomaly detection models is a challenging task due to the lack of anomaly-supervisory information and the unbounded nature of anomalies \cite{pang2021toward,ruff_unifying_2021}. Compare to the vast body of literature on the detection task, the research on interpretability in anomaly detection is very limited. There are a few existing works which focus on selecting discriminative features to explain the anomalous part of detected anomalies \cite{azmandian_local_2012,hutchison_local_2013,pang_unsupervised_2016,pang_sparse_2018}. Similar to feature selection, mining anomalous aspect such as finding the most outlying subspace of a given object \cite{duan_mining_2015,vinh_discovering_2016}, looking for numerical attributes \cite{angiulli_outlying_2017} and top feature pairs that best explain anomaly clusters \cite{liu_lp-explain_2020} have also been explored. Moreover, considering the intrinsic explanation ability of decision trees, some anomaly detectors adopt decision forest to extract explanation rules \cite{kopp_anomaly_2020,aguilar2022towards}. Besides, some deep anomaly detection models leverage attention mechanisms \cite{xu_beyond_2021}, feature-wise reconstruction errors \cite{schlegl_f-anogan_2019} and integrated gradients \cite{sipple_interpretable_2020} to provide interpretation for detected anomalies. There is also dependency-based method in which anomaly score is explained by the prediction errors of the features using other related features \cite{paulheim2015decomposition,feng2020relsen,lu2020lopad}. Compared with the existing works, our method can provide more straightforward information for understanding the reported anomalies to users even without a data science background.

Association rules have been used for anomaly detection since decades ago. Earlier examples include mining association rules for network intrusion detection \cite{mahoney2002learning}, credit card fraud detection~\cite{brause1999neural} and fault detection in spacecraft housekeeping data \cite{yairi2001fault}. More recently, association rules are used to detect anomalies in data collected from heating substations \cite{xue_fault_2017}, gas concentration equipment \cite{huang_research_2021}, soil moisture probes \cite{yu_automated_2018}, power systems \cite{kuang_association_2021} and software systems \cite{ge_apt_2021}, to name a few. A major advantage of association rule-based anomaly detection is its self-explainability, however, as association rules must be generated from a categorical dataset, their application is significantly restricted in practice. To overcome the problem, several methods are designed to transform continuous data to meaningful categorical data, e.g., via expert knowledge \cite{pal2017effectiveness} and auto-regressive models \cite{momtazpour2015analyzing}. Most similar to our work is \cite{feng2019systematic} in which tailored clustering and regression models are used to discretize sensor values for invariant rule mining in industrial control systems. However, the method is not general because some key assumptions made, e.g., value change of discrete variables (actuator states) is caused by some continuous variables (sensors) reaching critical values and the update of continuous variables follows a Gaussian distribution within a specific hidden control state, are only valid in industrial control systems. In this work, we leverage association rule mining and decision tree learning to provide a highly interpretable and general anomaly detection method of broader practical usage. Combining association rule mining and decision tree learning for interpretable anomaly detection has also been explored in \cite{kopp2015evaluation}. However, it is assumed that there exists a labeled anomaly dataset, and decision trees are trained to directly classify whether a data point is anomaly or not in \cite{kopp2015evaluation}. Different from \cite{kopp2015evaluation}, our method does not require such a labelled anomaly dataset which rarely exists in practice.

\section{Conclusion}
\label{sec:conclude}
We proposed a highly interpretable and general anomaly detection method by combining association rule mining and decision tree learning in a novel way to learn invariant rules from data. Since our learned invariant rules are self-explainable, they can provide straightforward clues to assist users to locate and understand the cause of the detected anomalies which is of vital importance in practice. Furthermore, experimental results on the benchmark datasets in various application domains show that our method can also achieve comparable or even better detection performance compared with several state-of-the-art anomaly detection algorithms. Like any study, our proposed method is not without limitation. First, our method currently only works for tabular data, we leave the extension of our method for other data types as future work. Second, our method works well for datasets with a considerable number of features, e.g., larger than 10. But its performance could be reduced if there are not enough features where we can derive enough predicates to generate invariant rules to cover the ``normal profile'' of the data generation process. Nevertheless, our method is a strong contribution to the anomaly detection community as providing tangible explanations of anomalies is imperative for the anomaly detection field \cite{ruff_unifying_2021}.
\bibliographystyle{ACM-Reference-Format}
\bibliography{kdd}


\end{document}